\documentclass[letterpaper, 10 pt, conference]{ieeeconf}  

\IEEEoverridecommandlockouts                              

\usepackage{multirow}
\usepackage{bbding}
\usepackage{pifont}
\usepackage[nointegrals]{wasysym}
\usepackage{caption}
\usepackage{subcaption}
\usepackage{amsmath}
\usepackage{array}
\usepackage{color}
\usepackage[normalem]{ulem}
\usepackage{hyperref}
\usepackage{nomencl}
\makenomenclature
\usepackage[ruled,vlined]{algorithm2e}
\usepackage{amsmath,amssymb,amsfonts}
\usepackage{algorithmic}
\usepackage{graphicx}
\usepackage{textcomp}
\usepackage{wrapfig}
\usepackage{mdframed,lipsum}
\usepackage[T1]{fontenc} 
\usepackage{booktabs}
\usepackage{mwe}
\usepackage[table, dvipsnames]{xcolor}
\usepackage{array,booktabs}

\usepackage{cite}
\def\BibTeX{{\rm B\kern-.05em{\sc i\kern-.025em b}\kern-.08em
    T\kern-.1667em\lower.7ex\hbox{E}\kern-.125emX}}
\usepackage{tikz}
\newcommand*{\tikzhl}[1]{\tikz[baseline=(X.base)] \node[fill=lightgray] (X) {#1};}

\usepackage{caption}
\title{\LARGE \bf
Learning Soft Robot Dynamics using Differentiable Kalman Filters and Spatio-Temporal Embeddings
}

\author{Xiao Liu$^{1}$, Shuhei Ikemoto$^{2}$, Yuhei Yoshimitsu$^{2}$, and Heni Ben Amor$^{1}$%
\thanks{$^{1}$X. Liu and H.~Ben Amor are with the School of Computing and Augmented Intelligence, Arizona State University
        {\tt\small \{xliu330, hbenamor\}@asu.edu}}
\thanks{$^{2}$S. Ikemoto and Y. Yoshimitsu are with Department of Human Intelligence Systems, Graduate School of Life Science and Systems Engineering, Kyushu Institute of Technology 
        {\tt\small ikemoto@brain.kyutech.ac.jp} and {\tt\small yoshimitsu.yuhei608@mail.kyutech.jp}}
}

\begin{document}

\maketitle
\thispagestyle{empty}
\pagestyle{empty}

\begin{abstract}

This paper introduces a novel approach for modeling the dynamics of soft robots, utilizing a differentiable filter architecture. The proposed approach enables end-to-end training to learn system dynamics, noise characteristics, and temporal behavior of the robot. A novel spatio-temporal embedding process is discussed to handle observations with varying sensor placements and sampling frequencies. The efficacy of this approach is demonstrated on a tensegrity robot arm by learning end-effector dynamics from demonstrations with complex bending motions. The model is proven to be robust against missing modalities, diverse sensor placement, and varying sampling rates. Additionally, the proposed framework is shown to identify physical interactions with humans during motion. The utilization of a differentiable filter presents a novel solution to the difficulties of modeling soft robot dynamics. Our approach shows substantial improvement in accuracy compared to state-of-the-art filtering methods, with at least a 24\% reduction in mean absolute error (MAE) observed. Furthermore, the predicted end-effector positions show an average MAE of 25.77mm from the ground truth, highlighting the advantage of our approach. The code is available at \url{https://github.com/ir-lab/soft_robot_DEnKF}.



\end{abstract}


\section{Introduction}
Soft robots are deformable structures that can be actuated and are composed of materials that form smooth curved shapes~\cite{lee2017soft}. Robots of this type have the ability to perform a large range of movements, including extension, contraction, bending, shearing, and twisting, which makes them highly adaptable to confined spaces~\cite{lee2020twister,yumbla2021human} such as in medical~\cite{burgner2015continuum} and industrial settings. Tensegrity structures, composed of compressive members that are supported by tensile cables~\cite{skeleton2009tensegrity}, have been utilized in the design of soft robots. Such tensegrity robots have become popular in recent years since they bridge the gap between an inherently flexible system and the ability to use rigid components~\cite{jung2018bio, kim2020rolling, ikemoto2021development}. Their design allows for effective resistance against compressive forces in specific directions while also maintaining overall flexibility. 
\begin{figure}[t!]
\centering
\includegraphics[width=\linewidth]{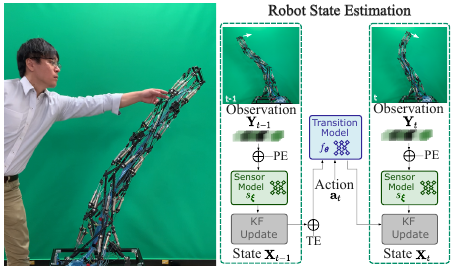}
\caption{{\bf Left}: A human is shown interacting with a soft robot; {\bf Right}: The differentiable Kalman filtering framework includes temporal and positional embedding modules (TE and PE) and a sensor model for projecting encoded raw observations to observation space. The filtering process involves utilizing the Kalman filter update step, where a stochastic transition model with an ensemble state is used to approximate the posterior distribution. }
\label{fig:overview}
  \vspace{-0.2in}
\end{figure} 

However, modeling the dynamics of soft robots with high-redundancy, flexibility, and a large number of degrees of freedom (DoF) is a daunting task due to their intricate, nonlinear design properties. 
Although numerical simulation tools~\cite{huang2020dynamic} based on discrete differential geometry have provided insightful ways to analyze robot dynamics, real robots are still widely modeled using analytical approaches, such as the finite element method~\cite{gong2018inverse} and Cosserat rod theory~\cite{renda2018unified}. However, these analytical methods also have limitations, such as the constraints imposed by limited computing resources and model generalizability. Different types of robots may require different mathematical modeling approaches. Moreover, a change to the properties of the robot, e.g., sensing or actuation, may require substantial human expertise and time investment to adjust the underlying model and its parameters.  
Soft robots also present modeling challenges due to uncertainties with regard to optimal number of sensors, sensor placement and temporal sampling frequency required to capture their dynamics accurately~\cite{mahoney2016inseparable}. More generally, soft robots are a manifestation of complex and highly non-linear systems which are known to be hard to model.

In this paper, we argue that recent innovations in modeling complex systems are particularly well-suited for learning and predicting the non-linear dynamics of soft robots. In particular, we focus on deep state-space models (DSSM)~\cite{NEURIPS2018_5cf68969}. These models learn to estimate states and measurements from observed sequences in a data-driven fashion~\cite{NEURIPS2018_5cf68969, klushyn2021latent, kloss2021train}. Accordingly, the use of DSSMs can help to overcome the aforementioned modeling challenges associated with soft robots. One set of algorithms based on DSSMs, namely Differentiable Filters (DFs), focuses on learning state transition and measurement models from data while preserving the mechanisms inherent to Bayesian recursive filtering. These properties render DFs ideal for systems with complex dynamics and sensor observations. In addition, DFs also provide interpretable state representations and uncertainty estimation techniques, which is crucial for safety-critical systems~\cite{lee2020multimodal}.


Extending prior works on DSSMs, this paper introduces a novel differentiable filter called differentiable Ensemble Kalman Filters (DEnKF), for modeling soft robots. It offers an end-to-end learning approach to estimate the state of the system, which is highly nonlinear and difficult to model analytically.  The main contributions are: 
\begin{itemize}
    \item The introduction of a positional embedding process that enables the spatial generalization of DEnKF by encoding observations with respect to their positions. As a result, learned models can account for changes to the location of sensors on the robot body.
    \item The use of a temporal embedding process that allows DEnKF to perform inference at variable rates and account for a multitude of time-delays due to sensing, hardware or communication.
    \item The modular structure of the framework separates the state dynamics from the end-to-end learning process, ensuring that the state remains dynamic even in the absence of observations.
    \item The paper also demonstrates a downstream application task of the framework for estimating human contact and physical interactions with the robot.
\end{itemize}

\section{Related work}
Soft robots are known to be difficult to model due to their highly nonlinear and often complex behavior~\cite{lee2017soft}. To address this challenge, previous works have proposed various modeling approaches with different assumptions. 

\textbf{Soft robot modeling}: A common approach for modeling soft robots is through steady-state models, which are equivalent to the kinematic model and assume that the system remains at rest in the absence of external forces. Steady-state models have been used in previous works, such as in~\cite{george2017learning}, but they have limitations in terms of reachability, efficiency, and speed of the controller. Another approach for modeling soft robots involves using the finite element method (FEM) to develop geometrically exact models, as proposed in~\cite{gong2018inverse}. FEM-based models have been shown to accurately capture the behavior of soft robots. However, despite recent advancements in FEM algorithm optimizations, real-time applications necessitate further advancements in hardware technology.
Data-driven methods, such as using multiple layer perceptrons to develop mappings from action space to state space, have also been proposed for soft robot modeling~\cite{george2017learning, jiang2017two}, but these methods have limitations in terms of generalizability.
Dynamic models, which account for the time-varying behavior of the system, have been shown to provide advantages for effective motion planning. For example, in~\cite{della2020model}, a model-based dynamic controller is proposed under the piecewise constant curvature assumption, and in~\cite{george2020first}, modeling based on a first-order dynamical system is used. However, due to the specificity of each dynamic model developed for different soft robot physics, it is challenging to achieve generalization.
Recently, a state estimation approach based on a sparse Gaussian process regression and Cosserat rod theory is proposed in~\cite{lilge2022continuum}, providing a general way of continuum robot modeling. However, this approach also has its own limitations, such as the need for high computational resources and the difficulty of incorporating prior knowledge of physical constraints. 

In general, each modeling approach is associated with certain advantages and limitations, and the selection of a particular approach hinges on diverse factors, including the particular application's demands, available resources, and the feasibility of sensor deployment.

\textbf{Differentiable filters}: Differentiable Filters (DFs) are composed of neural networks with algorithmic priors of Bayesian filter techniques to provide learning-based approaches for the forward and measurement models in recursive filtering. BackpropKF~\cite{haarnoja2016backprop} trains Kalman Filters using backpropagation with the integration of feed-forward networks and convolutional neural networks. Similarly, differentiable algorithm networks~\cite{karkus2019differentiable} introduce neural network components that encode differentiable robotic algorithms, akin to Differentiable Particle Filters (DPFs)~\cite{jonschkowski2018differentiable,chen2021differentiable}. DPFs employ algorithmic priors to increase learning efficiency and variations have been explored using adversarial methods for posterior estimation~\cite{wang2019dualsmc}. DFs were analyzed in~\cite{kloss2021train} for training and modeling uncertainty with noise profiles. The authors implemented the DFs as multi-layer perceptrons enveloped in an RNN layer, and tested them on real-world tasks in~\cite{kloss2021train,lee2020multimodal}. The results showed that end-to-end learning is crucial for accurately learning noise models.


\begin{figure}[t!]
\centering
\includegraphics[width=\linewidth]{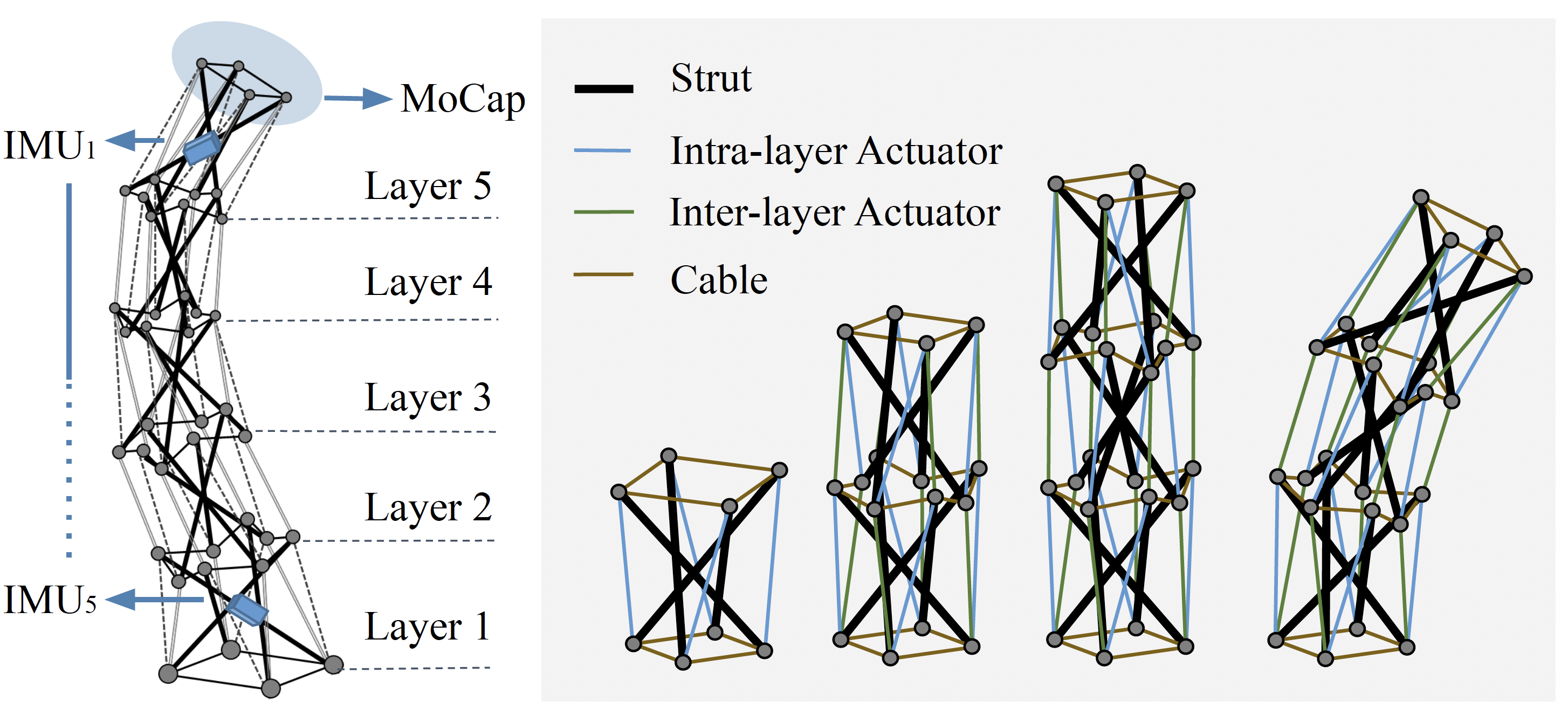}
\caption{The tensegrity robot: the robot contains 5 layers where each layer is a flexible tensegrity module with struts, stiff cables, and actuators. The sensory data associated with the robot are the IMUs, MoCap, and the pressure vector readings from the pneumatic cylinders actuators.}
\label{fig:structure}
  \vspace{-0.2in}
\end{figure} 

\section{Soft robot Modeling with DEnKF}
This section provides a detailed examination of the tensegrity robot structure, the bending motion mechanism, and the relevant sensory information. It then introduces the differentiable ensemble Kalman Filters (DEnKF) and its learning process as a method for propagating the robot state forward in time and correcting it using sensory data. To improve the framework's robustness and generalization across diverse sensor placements and inference rates, two enhancements -- Temporal and Positional embedding (TE and PE) -- are detailed to show how encoded state and observation features can be learned.

\subsection{Preliminaries}\label{Preliminaries}
The soft robot system employed in our work is a tensegrity robot arm developed in~\cite{ikemoto2021development}. This robot is built using a strict tensegrity structure, which consists of elements like struts, cables, spring-loaded cables, and actuated cables. The arm has five layers, as depicted in Fig.~\ref{fig:structure}. Each of these modules can be viewed as a layer of an arm-like tensegrity structure. By applying an external force, various continuous bending postures can be produced. As shown in Fig.~\ref{fig:structure}, the stiff cables maintain the longitudinal length while external forces contribute solely to changes in the bending direction. The sensory data associated with the robot, which is used to describe its kinematics, includes Inertial Measurement Unit (IMU) data, optical motion capture (MoCap) data, and proportional pressure control valves data. There are five IMUs mounted at each strut of the robot's layers, with each layer having four different struts for IMU placement. Thus, there are 20 different locations for IMU placement.

\begin{wrapfigure}{r}{0.20\textwidth}
\vspace{-0.2in}
  \begin{center}
    \includegraphics[width=0.19\textwidth]{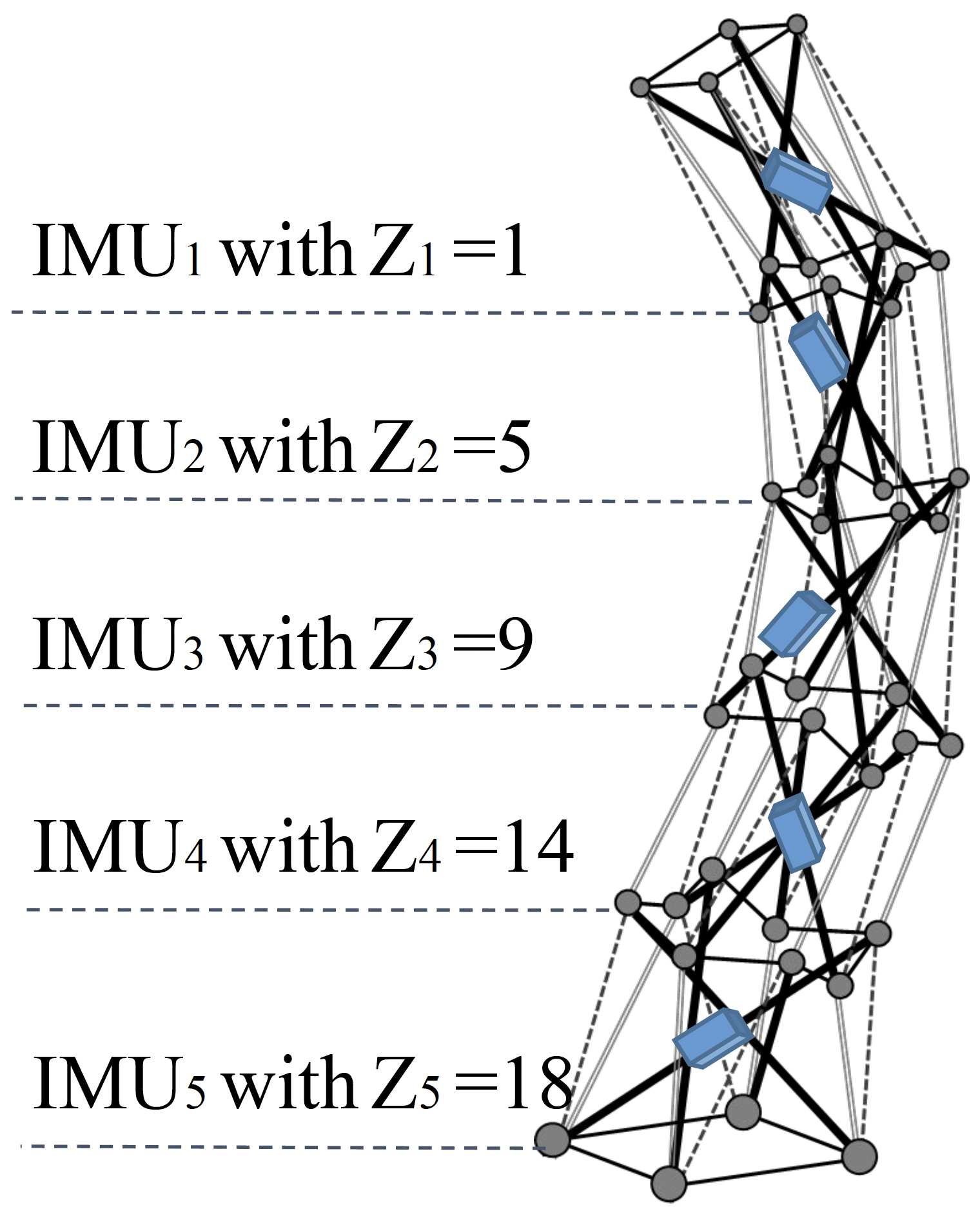}
  \end{center}
  \caption{The placement of the IMUs denoted by ${\bf{\mathcal{Z}}}$.}
  \label{fig:imu_position}
  \vspace{-0.1in}
\end{wrapfigure}
The state of a soft robot at time $t$ is represented by a 7-dimensional vector ${\bf x}_t = [x, y, z, {\bf q}_x, {\bf q}_y, {\bf q}_z, {\bf q}_w]^T$, which denotes the position and orientation of the robot's hand tip (end-effector).
The quaternion vector $\bf q$ represents the posture of the robot relative to the base frame (layer 1's bottom). 
The action ${\bf a}_t$ of the system is the pressure vector of the 40 pneumatic cylinder actuators, where ${\bf a}_t \in \mathbb{R}^{40}$. The raw observation ${\bf y}_t$ consists of 5 IMU readings, where ${\bf y}_t \in \mathbb{R}^{30}$, with each IMU providing a 6-dimensional vector of accelerations and angular velocities relative to its location. An example of the IMU placements is shown in Fig.~\ref{fig:imu_position}, where the blue cubes represent the location of the IMUs on each strut at each layer. Each integer number ${\bf{\mathcal{Z}}}_i$ denotes a location label, and a vector of integers ${\bf{\mathcal{Z}}}$ is recorded for five IMU locations, i.e., ${\bf{\mathcal{Z}}} = [1, 5, 9, 14, 18]$, with $1 \leq {\bf{\mathcal{Z}}}_i \leq 20$.

\subsection{Bayesian Filtering}
Recursive Bayesian filtering is a technique used to estimate the state ${\bf x}_t$ of a discrete-time dynamical system, given a sequence of actions ${\bf a}_{1:t}$ and noisy observations ${\bf y}_{1:t}$. The posterior distribution of the state can be represented by the following equation:
\begin{equation}
\begin{aligned}
\label{eq:01}
    &p({\bf x}_t | {\bf a}_{1:t}, {\bf y}_{1:t}, {\bf x}_{1:t-1}) \\
    &\propto
    p({\bf y}_{t} | {\bf a}_{t}, {\bf x}_t)\ p({\bf x}_t | {\bf a}_{1:t-1}, {\bf y}_{1:t-1}, {\bf x}_{1:t-1}).
\end{aligned}
\end{equation}
We can denote the belief of the state as $\text{bel}({\bf x}_t) = p({\bf x}_t | {\bf a}_{1:t}, {\bf y}_{1:t}, {\bf x}_{1:t-1})$. Assuming the Markov property, where the next state is dependent only on the current state, we get the following expression:
\begin{equation}
\begin{aligned}
\label{eq:02}
    \text{bel}({\bf x}_t) = \underbrace{p({\bf y}_{t} | {\bf x}_t)}_{\text{observation model}}
    \prod_{t=1}^t \overbrace{ p({\bf x}_t|{\bf a}_{t}, {\bf x}_{t-1})}^{\text{state transition model}} \text{bel}({\bf x}_{t-1}),
\end{aligned}
\end{equation}
where $p({\bf y}_{t} | {\bf x}_t)$ is the observation model and $p({\bf x}_t|{\bf a}_{t},{\bf x}_{t-1}$) is the transition model. The transition model describes the laws that govern the evolution of the system state, while the observation model identifies the relationship between the hidden, internal state of the system and observed, noisy measurements.

In our work, we introduce a method for state estimation called differentiable ensemble Kalman Filters (DEnKF). Our approach combines the traditional EnKF algorithm with recent advancements in stochastic neural networks (SNNs)~\cite{gal2016dropout} who established a connection between Dropout training and Bayesian inference in deep Gaussian processes. The EnKF algorithm involves updating the approximate posterior distribution by propagating each ensemble member forward in time. In DEnKF approach, we maintain the core algorithmic steps of EnKF while leveraging the capabilities of SNNs. To represent the initial state distribution, we use an ensemble of $E$ members, denoted by ${\bf X}_0 = [ {\bf x}^{1}_0, \dots, {\bf x}^{E}_0]$, where $E \in \mathbb{Z}^+$. The key difference between our approach and traditional EnKF is the implicit modeling of process noise through sampling from a stochastic neural network.

\textbf{Prediction Step}: We leverage the stochastic forward passes from a trained state transition model to update each ensemble member: 
    \begin{equation}
    \begin{aligned}\label{eq:1}
          {\bf x}^{i}_{t|t-1} & \thicksim  f_{\pmb {\theta}} ({\bf x}^{i}_{t|t-1}|{\bf a}_{t}, {\bf x}^{i}_{t-1|t-1}), \  \forall i \in E.
    \end{aligned}
   \end{equation}
 Matrix ${\bf X}_{t|t-1} = [{\bf x}^{1}_{t|t-1}, \cdots, {\bf x}^{E}_{t|t-1}]$ holds the updated ensemble members which are propagated one step forward through the state space. Note that sampling from the transition model $f_{\pmb {\theta}}(\cdot)$ (using the SNN methodology described above) implicitly introduces a process noise.

\textbf{Update Step}: Given the updated ensemble members ${\bf X}_{t|t-1}$, a nonlinear observation model $h_{\pmb {\psi}}(\cdot)$ is applied to transform the ensemble members from the state space to observation space. Following our main rationale, the observation model is realized via a neural network with weights $\pmb {\psi}$. Accordingly, the update equations for the EnKF become:
    \begin{align}
    \label{eq:2}
        {\bf H}_t {\bf X}_{t|t-1} &= \left[ h_{\pmb {\psi}}({\bf x}^1_{t|t-1}), \cdots, h_{\pmb {\psi}}({\bf x}^E_{t|t-1}) \right],\\
        \label{eq:3}
        {\bf H}_t {\bf A}_{t} &=  {\bf H}_t {\bf X}_{t|t-1} \\
        &- \left[\frac{1}{E} \sum_{i=1}^E h_{\pmb {\psi}}({\bf x}^i_{t|t-1}),
        \cdots,
        \frac{1}{E} \sum_{i=1}^E h_{\pmb {\psi}}({\bf x}^i_{t|t-1})\right]. \nonumber
    \end{align}
${\bf H}_t {\bf X}_{t|t-1}$ is the predicted observation, and ${\bf H}_t {\bf A}_{t}$ is the sample mean of the predicted observation at $t$. EnKF treats observations as random variables. Hence, the ensemble can incorporate a measurement perturbed by a small stochastic noise thereby accurately reflecting the error covariance of the best state estimate~\cite{evensen2003ensemble}. In our differentiable version of the EnKF, we also incorporate a sensor model which can learn projections between a latent space and observation space. We train a stochastic sensor model $s_{\pmb {\xi}}(\cdot)$:
    \begin{equation}
    \begin{aligned}\label{eq:sensor}
          \tilde{{\bf y}}^{i}_t & \thicksim  s_{\pmb {\xi}} (\tilde{{\bf y}}^{i}_t|{\bf y}_{t}),\  \forall i \in E,
    \end{aligned}
   \end{equation}
where ${\bf y}_{t}$ represents the raw observation. Sampling yields learned observations $\tilde{{\bf Y}}_t = [\tilde{{\bf y}}^{1}_t, \cdots, \tilde{{\bf y}}^{E}_t]$ and the sample mean $\tilde{{\bf y}}_t = \frac{1}{E}\sum_{i=1}^i\tilde{{\bf y}}^i_t$.  The innovation covariance ${\bf S}_t$ can then be calculated as:
    \begin{equation}
    \begin{aligned}\label{eq:4}
        {\bf S}_t &= \frac{1}{E-1}  ({\bf H}_t {\bf A}_t)  ({\bf H}_t {\bf A}_t)^T + r_{\pmb {\zeta}}(\tilde{{\bf y}_t}),
    \end{aligned}
    \end{equation}
where $r_{\pmb {\zeta}}(\cdot)$ is the measurement noise model implemented using multi-layer perceptron (MLP), it takes a learned observation $\tilde{{\bf y}_t}$ at time $t$ and provides a stochastic noise in the observation space by constructing the diagonal of the noise covariance matrix. The final estimate of the ensemble ${\bf X}_{t|t}$ can be obtained by performing the KF update step:
    \begin{align}
    \begin{split}\label{eq:5}
        {\bf A}_t &= {\bf X}_{t|t-1} - \frac{1}{E}\sum_{i=1}^E{\bf x}^i_{t|t-1},
    \end{split}\\
    \begin{split}\label{eq:6}
     {\bf K}_t &= \frac{1}{E-1} {\bf A}_t ({\bf H}_t {\bf A}_t)^T {\bf S}_t^{-1},
    \end{split}\\
    \begin{split}\label{eq:7}
    {\bf X}_{t|t} &= {\bf X}_{t|t-1} + {\bf K}_t (\tilde{{\bf Y}}_t - {\bf H}_t {\bf X}_{t|t-1}),
    \end{split}
    \end{align}
where ${\bf K}_t$ is the Kalman gain. In inference, the ensemble mean ${\bf \bar{x}}_{t|t} = \frac{1}{E}\sum_{i=1}^E {\bf x}^i_{t|t}$ is used as the updated state. The neural network structures for sub-modules are described in Table~\ref{tab:EnKF_module}.

\begin{table}[t!]
  \centering
  \caption{Differentiable filters' learnable sub-modules.}
  \label{tab:EnKF_module}
  \scalebox{0.90}{
  \begin{tabular}{ll}
    \toprule
$f_{\pmb {\theta}}$: & 2$\times$SNN(64, ReLu), 2$\times$SNN(128, ReLu), 1$\times$SNN(S, -)\\
$h_{\pmb {\psi}}$: & 2$\times$fc(32, Relu), 2$\times$fc(64, ReLu), 1$\times$ fc(O, -)\\
$r_{\pmb {\zeta}}$: & 2$\times$fc(16, ReLu), 1$\times$fc(O, -)\\


\multirow{2}{1em}{$s_{\pmb {\xi}}$:} & 
 fc(128, ReLu), flatten(), 2$\times$SNN(512, ReLu), 1$\times$SNN(256, ReLu), \\
 & 1$\times$SNN(128, ReLu), 1$\times$SNN(O, -)\\
    \bottomrule
\multicolumn{2}{l}{fc: fully connected, conv: convolution, S, O: state and observation dimension.} \\
  \end{tabular}}
\end{table}

\subsection{Spatio-Temporal Embedding}
\begin{figure}[t!]
\centering
\includegraphics[width=\linewidth]{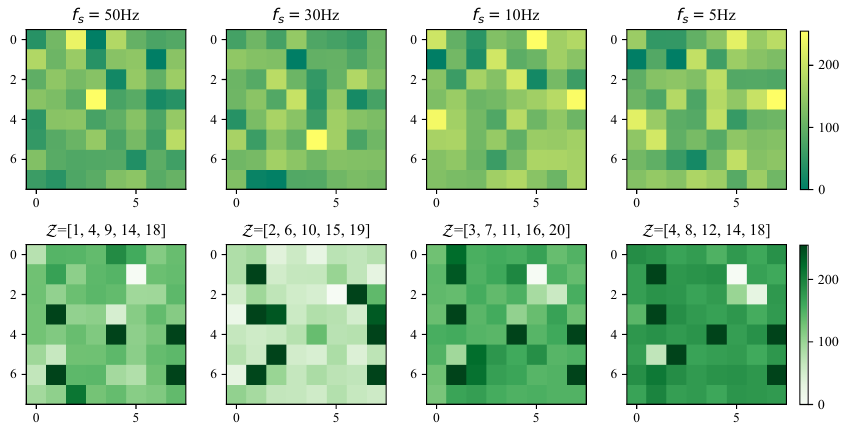}
\caption{The top and bottom figures show the output latent feature maps from the temporal embedding process and the positional embedding process, respectively, for various sampling frequencies $f_s$ and IMU positions ${\bf{\mathcal{Z}}}$. }
\label{fig:PE}
  \vspace{-0.2in}
\end{figure} 
The aim of the temporal and positional embedding is to increase the generalizability of the robot state estimator under varied inference frequencies and sensor positions. Taking inspiration from language models, which encode different positional information of words in a sentence~\cite{vaswani2017attention}, we incorporate temporal and positional labels in a similar fashion. We opt to use sinusoidal functions implemented in~\cite{vaswani2017attention} with varying frequencies using the temporal and positional embedding functions:
    \begin{equation}
    \begin{aligned}
        \text{PE}_{(pos, i)} = \left\{ \begin{array}{rcl}
            \sin(\frac{pos}{10000^{2i/ d_{m}}}) & \mbox{if}
            \ i \ \text{mod} \ 2 = 0 \\ 
            \cos(\frac{pos}{10000^{2i/ d_{m}}}) & \mbox{otherwise}\\
            \end{array}\right.
    \end{aligned}
    \end{equation}
where $pos$ is the position and $i$ is the dimension, $d_m$ is the dimension of the feature space. 
The embedding function is first passed the different sensor placements denoted as ${\bf{\mathcal{Z}}}$, and then the sensor model takes the embedded input features from the observation:
    \begin{equation}
    \begin{aligned}
          \tilde{{\bf y}}^{i}_t & \thicksim  s_{\pmb {\xi}} (\tilde{{\bf y}}^{i}_t|  {\bf y}_{t}, \tikzhl{ ${\bf{\mathcal{Z}}}$ }),\  \forall i \in E.
    \end{aligned}
   \end{equation}
The latent feature maps displayed in Fig.~\ref{fig:PE} exhibit varied encodings of ${\bf{\mathcal{Z}}}$, which enables $s_{\pmb{\xi}}$ to learn from the distinctive features presented in each encoding. The temporal embedding process is used when the system progresses at different speeds. In our system, we offer four options for the sampling frequency $f_s$ -- 5Hz, 10Hz, 30Hz, and 50Hz. The temporal embedding process is integrated within the latent space of the state transition model as follows:
    \begin{equation}
    \begin{aligned}
          {\bf x}^{i}_{t|t-1} & \thicksim  f_{\pmb {\theta}} ({\bf x}^{i}_{t|t-1}|{\bf a}_{t}, {\bf x}^{i}_{t-1|t-1},   \tikzhl{$f_s$}), \  \forall i \in E.
    \end{aligned}
   \end{equation}
Here, the input state ${\bf X}_{t-1}$ along with its corresponding $f_s$ are encoded by $f_{\pmb{\theta}}(\cdot)$ into a 64-dimensional latent vector, which is then used to apply the state transition. The encoded latent vector from the same initial state ${\bf X}_{0}$ is shown in Fig.~\ref{fig:PE} for different values of $f_s$. By doing so, $f_{\pmb{\theta}}(\cdot)$ is able to learn the transitions from the unique state features.

\section{Experiment}
This section presents a series of experiments that are conducted to evaluate the performance of the proposed differentiable ensemble Kalman filters (DEnKF) for state tracking. A comparison with baseline differentiable filters~\cite{kloss2021train,jonschkowski2018differentiable} is also discussed. Furthermore, two downstream tasks of state estimation are performed, namely estimation with missing observations and virtual force estimation.

\begin{table*}[h!]
\caption{Ablation study of proposed DEnKF with and without positional and temporal embedding processes for state estimation task in different conditions. The MAE error metric of the 10-fold cross-validation is reported.}
\label{Tab:ablation}
\begin{center}
\scalebox{1}{
\begin{tabular}{l c c c c c c }
    \toprule
     \multirow{2}{3em}{Method} 
     &\multicolumn{2}{c}{Fixed ${\bf{\mathcal{Z}}}$}
     &\multicolumn{2}{c}{Multiple ${\bf{\mathcal{Z}}}$}
     &\multicolumn{2}{c}{Multiple ${\bf{\mathcal{Z}}}$ and $f_s$}
     \\
     &EE (mm)  &$\bf q$
     &EE (mm)  &$\bf q$
     &EE (mm)  &$\bf q$
     \\
    \midrule
     DEnKF-Fix
     &\bf 25.7765$\pm$7.827
     &0.0648$\pm$0.035
     &51.7728$\pm$7.489
     &0.1996$\pm$0.034
     &106.2310$\pm$17.760
     &0.1722$\pm$0.065
     \\
    DEnKF-PE
     &29.6578$\pm$8.873
     &\bf 0.0626$\pm$0.028
     &22.5427$\pm$9.146
     &0.0768$\pm$0.033
     &71.5244$\pm$9.411
     &0.1844$\pm$0.022
     \\
    DEnKF-PE+TE
     &31.5985$\pm$9.582
     &0.0788$\pm$0.021
     &\bf 21.7658$\pm$6.337
     &\bf 0.0641$\pm$0.026
     &\bf 25.7566$\pm$4.835
     &\bf 0.0466$\pm$0.055
     \\
\bottomrule
\multicolumn{7}{l}{Means$\pm$standard errors.} \\
\end{tabular}}
\end{center}
\vspace{-0.1in}
\end{table*}

\subsection{Experimental Setup}
\textbf{Data}: The dataset is obtained by performing optical motion capture on the real tensegrity robot hand tip while supplying randomly generated desired pressure vectors to the pneumatic cylinder actuators. As mentioned in Sec.~\ref{Preliminaries}, we record 40-dimensional pressure vectors as the action ${\bf a}_t \in \mathbb{R}^{40}$, 5 IMU readings ${\bf y}_t \in \mathbb{R}^{30}$ with the corresponding position ${\bf{\mathcal{Z}}}$, and a 7-dimensional state ${\bf x}_t$. We collect 10 time-series data $D_1 - D_{10}$ (shown as Table~\ref{tab:position}), with each continuous sampling lasting for one hour. There are 12,000 trials collected in total. During each trial, the robot is moved from the current equilibrium posture to the next equilibrium posture by applying the new desired pressure. We down-sample each time-series data to obtain the dataset with different sampling frequencies $f_s$. All data is gathered through the a ROS2 network, and their synchronization is achieved using the ``message\_filters" package. 

\begin{table}[h!]
  \centering
  \caption{Time-series data with different sensor placements.}
  \label{tab:position}
  \scalebox{1}{
  \begin{tabular}{l}
    \toprule
$D_1$ : [1,4,9,14,18],  $\ D_2$ : [1,5,9,15,19],  $\ D_3$ : [2,6,10,15,19] \\
$D_4$ : [2,6,10,16,20],  $D_5$ : [2,6,10,13,17], $D_5$ : [3,7,11,13,17]\\
$D_6$ : [3,7,11,14,18],  $D_7$ : [3,7,11,16,20], $D_8$ : [4,8,12,16,20]\\
$D_9$ : [4,8,12,14,18],  $D_{10}$ : [4,8,12,15,19]\\
    \bottomrule
  \end{tabular}}
  \vspace{-0.2in}
\end{table}

\begin{figure*}[h]
\centering
\includegraphics[width=0.95\linewidth]{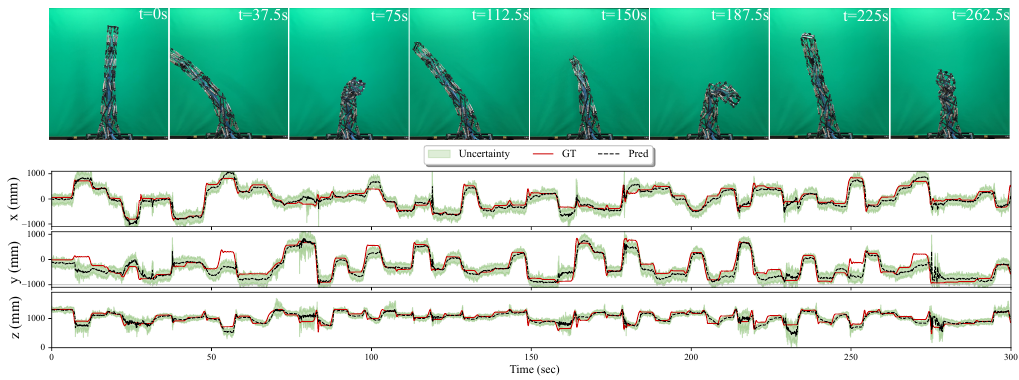}
\caption{The real-time estimation of the state on the tensegrity robot arm is demonstrated in this study. Specifically, the {\bf top} figure presents a motion sequence of the robot without applying external forces, while the {\bf bottom} figure showcases the real-time tracking outcomes (along with the corresponding uncertainty) of the positions of the hand tip.}
\label{fig:real_robot}
  \vspace{-0.2in}
\end{figure*}

\textbf{Training}: Proposed framework contains four sub-modules as listed in Table~\ref{tab:EnKF_module}. The entire framework is trained in an end-to-end manner via a mean squared error (MSE) loss between the ground truth state $\hat{{\bf x}}_{t|t}$ and the estimated state ${\bf \bar{x}}_{t|t}$ at every timestep. We also supervise the intermediate modules via loss gradients $\mathcal{L}_{f_{\pmb {\theta}}}$ and $\mathcal{L}_{s_{\pmb {\xi}}}$. Given ground truth at time $t$, we apply the MSE loss gradient calculated between $\hat{{\bf x}}_{t|t}$ and the output of the state transition model to $f_{\pmb {\theta}}$ as in Eq.~\ref{eq:loss1}. We apply the intermediate loss gradients computed based on the ground truth observation $\hat{{\bf y}_t}$ and the output of the stochastic sensor model $\tilde{{\bf y}}_t$: 
    \begin{align}
    \label{eq:loss1}
    \mathcal{L}_{f_{\pmb {\theta}}} =  \| {\bf \bar{x}}_{t|t-1} - \hat{{\bf x}}_{t|t}\|_2^2,\ \ 
        \mathcal{L}_{s_{\pmb {\xi}}} =\| \tilde{{\bf y}_t} -  \hat{{\bf y}_t}\|_2^2.
    \end{align}
All models in the experiments were trained for 50 epochs with batch size 64, and a learning rate of $\eta = 10^{-5}$. The ensemble size of the Kalman filter was set to 32 ensemble members.

\subsection{State Estimation}
In this experiment, we investigate the effectiveness of proposed differentiable filters in estimating the state of the tensegrity robot, and perform comparisons against other differentiable filters baselines. To reiterate, the robot state is defined as ${\bf x}_t = [x, y, z, {\bf q}_x, {\bf q}_y, {\bf q}_z, {\bf q}_w]^T$, the learned observation $\tilde{{\bf y}}_t$ is defined to have the same dimension as the robot state, where $\tilde{{\bf y}}_t = [x, y, z, {\bf q}_x, {\bf q}_y, {\bf q}_z, {\bf q}_w]^T$. The state estimator tracks the robot end-effector (EE) in position and orientation while random generated pressure vectors are supplied for the pneumatic cylinder actuators.

\textbf{Results}: The proposed DEnKF leverages both the positional embedding (PE) and temporal embedding (TE) processes. We conducted a comprehensive evaluation of the performance of DEnKF under different ${\bf{\mathcal{Z}}}$ conditions and with varying $f_s$. The experiment involved training and validating three models: a) DEnKF-Fix trained on a single time-series data, b) DEnKF-PE trained with multiple time-series data collected from diverse ${\bf{\mathcal{Z}}}$ conditions, and c) DEnKF-PE+TE trained with multiple time-series data from various ${\bf{\mathcal{Z}}}$ conditions and $f_s$ values. Table~\ref{Tab:ablation} presents the ablation study results for the three models evaluated with 10-fold cross-validation and the mean absolute error (MAE) metric. In the ablation study, three different conditions are considered -- the fixed IMU positions, multiple IMU positions, and multiple IMU positions with various sampling frequencies. The DEnKF-Fix model, trained with a single time-series data and a fixed IMU placement, shows the best performance with an average 25.77mm offset from the ground truth EE positions. However, its performance is limited to fixed IMU placement only. The DEnKF-PE+TE model demonstrates robust performance across different ${\bf{\mathcal{Z}}}$ conditions and improves the accuracy of the EE positions and orientations by 58\% and 66\% respectively, compared to the DEnKF-Fix model. While DEnKF-PE achieves comparable MAE to DEnKF-PE+TE under multiple ${\bf{\mathcal{Z}}}$ conditions, the performance of DEnKF-PE+TE outperforms DEnKF-PE by 64\% and 75\% on position and orientation respectively, when multiple $f_s$ values are provided.

\begin{figure}[h]
\centering
\includegraphics[width=0.92\linewidth]{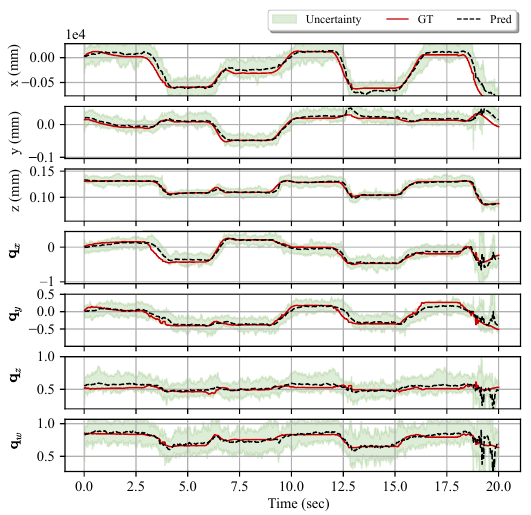}
\caption{State estimation results on the EE positions ($x, y, z$) and orientations presented as $({\bf q}_x, {\bf q}_y, {\bf q}_z, {\bf q}_w)$.}
\label{fig:result_full_state}
\vspace{-0.1in}
\end{figure}

\begin{figure}[t!]
\centering
\includegraphics[width=0.8\linewidth]{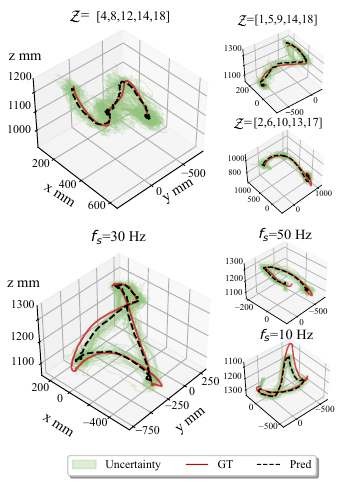}
\caption{Testing results on state estimation with various IMU positions ${\bf{\mathcal{Z}}}$ and sampling frequencies $f_s$.}
\label{fig:result_state}
\vspace{-0.1in}
\end{figure}
To perform a comprehensive evaluation of the DEnKF framework, a long-term state estimation experiment lasting for five minutes was carried out on the physical tensegrity robot, as depicted in Fig.~\ref{fig:real_robot}. The video sequence captures the robot in different configurations at various points in time, while the estimated positions of the EE demonstrate robust and stable state tracking throughout the entire duration of the experiment. Moreover, the complete 7-dimensional state vector, which includes both position and orientation, is shown in Fig.~\ref{fig:result_full_state}. While the orientation vector exhibits a relatively larger uncertainty, the framework is able to accurately track the overall state over time. Further analysis of the state tracking is conducted by visualizing the EE trajectories in 3D for the DEnKF-PE+TE model, as shown in Fig.~\ref{fig:result_state}. Each trajectory represents one test trial, and the prediction results of the DEnKF model are shown with the ensemble state outputs representing the uncertainty. The trajectories show the robustness and adaptability of DEnKF across a range of conditions, including multiple distinct ${\bf{\mathcal{Z}}}$ and $f_s$.

\begin{table}[h!]
\caption{Comparison with other baselines on state estimation task measured in RMSE and MAE of the EE position with fixed IMU locations ${\bf{\mathcal{Z}}}$. Results for dEKF, dPF, and dPF-M-lrn are reproduced for detailed comparisons.}
\label{Tab:comparison}
\begin{center}
\scalebox{1}{
\begin{tabular}{c c c c }
    \toprule
     \multirow{2}{3em}{Method} 
     &\multicolumn{2}{c}{Fixed IMU ${\bf{\mathcal{Z}}}$}
     &\multirow{2}{5em}{Wall clock time (s)}
     \\
     &RMSE  &MAE
     &
     \\
    \midrule
     dEKF~\cite{kloss2021train}
     & 61.753$\pm$1.630 
     & 41.960$\pm$1.147
     &\bf 0.047
     \\
    DPF~\cite{jonschkowski2018differentiable}
     &51.184$\pm$7.204
     &34.317$\pm$4.205
     &0.060
     \\
    dPF-M-lrn~\cite{kloss2021train}
     &49.799$\pm$8.264
     &33.903$\pm$6.964
     &0.059
     \\
    DEnKF-Fix (ours)
     &\bf 31.519$\pm$9.974
     &\bf 25.777$\pm$7.827
     &0.062
     \\
\bottomrule
\multicolumn{4}{l}{Means$\pm$standard errors.} \\
\end{tabular}}
\end{center}
\vspace{-0.2in}
\end{table}

\textbf{Comparison}: Table~\ref{Tab:comparison} presents the performance comparison of our proposed differentiable Kalman filter with fixed sensor model (DEnKF-Fix) against state-of-the-art differentiable filters for state estimation of soft robots, namely differentiable Extended Kalman Filters (dEKF)\cite{kloss2021train}, differentiable Particle Filters (DPF)\cite{jonschkowski2018differentiable}, and the modified differentiable Particle Filter with learned process noise model (dPF-M-lrn)~\cite{kloss2021train}. To ensure a fair comparison, we removed the temporal and positional embedding processes (TE and PE) for the DEnKF and supplied the same sensor model ${s_{\pmb {\xi}}}$ for all methods. For the DPF and dPF-M-lrn methods, we trained and tested with 100 particles.
 Our results indicate that the DEnKF-Fix approach outperforms the state-of-the-art methods with a Mean Absolute Error (MAE) of 25.78mm. Specifically, our approach achieved an MAE that is 24\%, 25\%, and 39\% lower than that of dEKF, DPF, and dPF-M-lrn, respectively. Among the baselines, dPF-M-lrn shows slightly better results than others, it does not exhibit any advantages over DEnKF-Fix. Our findings highlight the effectiveness of the DEnKF for state estimation of soft robots, particularly in comparison to existing differentiable filter approaches.

\subsection{State Estimation with Missing Observation}
In the field of soft robotics, sensor failures are common and can be mitigated by using learning-based sensing techniques~\cite{ang2022learning}. The modular structure of the proposed framework offers an additional advantage by enabling compensation for such issues. In this experiment, we investigate the robustness of DEnKF in the event of sensor failures. Specifically, we use the forward model ${f_{\pmb {\theta}}}$ to update the robot state in the absence of observations. The experiment is conducted on a trained DEnKF-PE+TE model, which has not been exposed to such scenarios during training. 

\begin{figure}[t!]
\centering
\includegraphics[width=\linewidth]{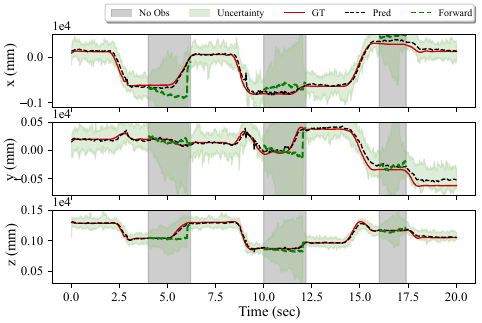}
\caption{State estimation results with missing observations (grey) using DEnKF; Only the forward model $f_{\pmb {\theta}}$ is used when no observation is obtained.}
\label{fig:forward}
  \vspace{-0.2in}
\end{figure}

\textbf{Results}: In the experiment, we enable random 12.5\% and 6.26\% time windows with no observations for each testing trial (with 20 seconds). Figure~\ref{fig:forward} demonstrates one of the results when the state estimator is in the scenario where no observations are available. The prediction results show the state tracking outcomes with observation, while the green line represents the case where the forward model ${f_{\pmb {\theta}}}$ alone is used. The uncertainty, which is described by the distribution of the ensemble members, increases when no observations are obtained and decreases when observations are obtained again.

\begin{figure}[h!]
\centering
\includegraphics[width=\linewidth]{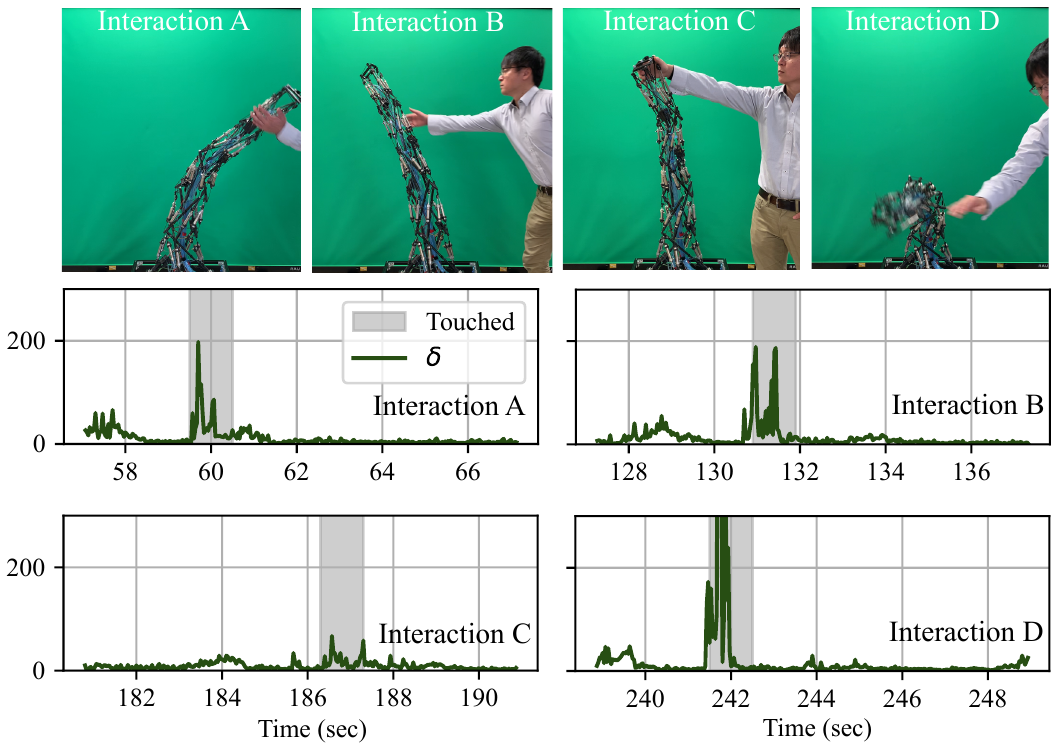}
\caption{Experiment on estimating virtual forces: in a testing sequence of 5 minutes, the robot is subjected to different forces labeled as A, B, C, and D, the virtual forces $\delta$ is captured during each interaction.}
\label{fig:force}
  \vspace{-0.2in}
\end{figure} 
\subsection{Virtual Force Estimation}
An important downstream application of DEnKF is virtual force estimation. During training, the forward model ${f_{\pmb{\theta}}}$ learns the system dynamics with given actions and previous states in the prediction step of DEnKF. However, in the correction step, which involves the sensor model $s_{\pmb {\xi}}$ and Kalman update, only a time-invariant mapping between observation space and state space is conducted, without considering the state dynamics. In other words, the correction step estimates the state based on the posture configurations of the robot. Therefore, the correction step always generates a ``corrected" state ${\bf \bar{x}}_{t|t}$ at time $t$ regardless of whether a force is applied to the robot or not, but the prediction step outputs an estimated state $\hat{{\bf x}}_{t|t-1}$ assuming no force is applied. The virtual force defined as $\delta  = \| {\bf \bar{x}}_{t|t} - \hat{{\bf x}}_{t|t-1} \|_p$, it is estimated between these two different outputs using the Minkowski distance:
    \begin{align}
    \label{eq:distance}
    \| {\bf \bar{x}}_{t|t} - \hat{{\bf x}}_{t|t-1} \|_p = \left(\sum | {\bf \bar{x}}_{t|t} - \hat{{\bf x}}_{t|t-1}|^p\right)^{\frac{1}{p}}
    \end{align}
with $p=10$.

\textbf{Results}: In this experiment, we apply random pressure vectors to the pneumatic cylinder actuators and conduct several interactions with the robot by applying external forces of different directions and magnitudes. The forces are applied four times, labeled A, B, C, and D, during a test sequence of 5 minutes as shown in Fig.~\ref{fig:force}. We capture the virtual force $\delta$ value during testing, also shown in Fig.~\ref{fig:force}. The gray areas indicate the time window of interaction. It is apparent that $\delta$ reflects the forces during each interaction. Interestingly, the magnitude of the Minkowski distance $\delta$ is proportional to the actual forces applied to the robot. For instance, in interaction D, where the actual forces are significant enough to entirely change the posture configurations of the robot (snapshot D in Fig.~\ref{fig:force}), the $\delta$ value reaches a higher value (>400) compared to the other interactions.

\section{Conclusions}
This paper presents a comprehensive study on the modeling of soft robot dynamics using differentiable ensemble Kalman filters (DEnKF). To enhance the spatio-temporal generalizability of state estimation, the proposed approach integrates temporal and positional embedding (TE and PE) processes. Through experiments on a highly nonlinear system, specifically the tensegrity robot arm, the proposed DEnKF demonstrates stable and accurate state tracking. In comparison to other differentiable filter frameworks, the proposed DEnKF outperforms the baselines and is capable of handling other downstream tasks, such as missing observation and virtual force estimation. Notably, the proposed DEnKF approach allows for fine-grained analysis of the state forward model and sensor model, which is not supported by other baselines. Because those baselines are RNN-based filters and the forward model has to remain in the same hidden state until an observation is processed.

With respect to limitations, this study focuses exclusively on the application of IMU sensors, while other types of soft robot sensors, such as fiber-based deformation sensors~\cite{liu2015large} and tendon/backbone strain sensors~\cite{xu2008investigation}, have not been tested. It is important to acknowledge that each sensor has unique characteristics, and a fusion module may be able to leverage the advantages of multiple sensors based on the specific application requirements.
Regarding force estimation, while the Minkowski distance has demonstrated the ability to reflect actual external forces, it is important to note that actual forces can also be learned and calibrated using supervised learning techniques. Additionally, other distance metrics can be explored to find the best fit for a particular application. Therefore, further research is needed to investigate the potential benefits and limitations of different types of soft robot sensors and distance metrics for force estimation.

In future research, the authors aim to extend the functionality of the DEnKF framework by exploring the calibration of virtual forces to actual values for improved force estimation. Additionally, the authors plan to investigate alternative state representations, such as segment-wise estimation of the soft robot, which may facilitate a more detailed understanding of the deformation characteristics of the robot and enhance estimation accuracy. These proposed efforts are expected to enhance the versatility and efficacy of the DEnKF framework and further advance its applicability in the domain of soft robotics.

\section*{Acknowledgement}
This research was funded by the JSPS Grants-in-Aid for Scientific Research (KAKENHI) program under grant numbers 19K0285, 19H01122, and 21H03524. This research was also funded partially by ``The Global KAITEKI Center” (TGKC) of the Global Futures Laboratory at Arizona State University.

\bibliographystyle{IEEEtran}
\bibliography{IEEEabrv,reference}

\end{document}